\documentclass[journal, letterpaper]{IEEEtran}

\usepackage{graphicx}
\usepackage{url}         
\usepackage{textgreek}	
\usepackage{listings}
\usepackage{csvsimple}
\usepackage{longtable}

\usepackage[colorlinks=true, allcolors=blue]{hyperref}

\usepackage{natbib}  
\usepackage{caption} 

\usepackage{algorithm}
\usepackage{algorithmic}
\usepackage{booktabs}
\usepackage{amsmath}
\usepackage{amssymb}
\usepackage{multirow}
\usepackage{authblk}
\usepackage{pifont}
\newcommand{\dvis}{\delta_{\text{vis}}}
\newcommand{\dprior}{\delta_{\text{prior}}}
\newcommand{\snr}{\text{SNR}}
\newcommand{\FPR}{\text{FPR}}

\title{When Visual Signals Mislead: A Mechanistic Study of 
Attribute Hallucination in Vision-Language Models}

\author[1]{Yufei Zhang}
\author[1]{Chenlu Zhan}
\author[1]{Hongwei Wang*}

\affil[1]{Zhejiang University}

\begin{document}
\maketitle

\begin{abstract}
Attribute hallucination---where vision-language models (VLMs) correctly
identify an object but mischaracterize its properties---is prevalent yet
mechanistically poorly understood. The dominant explanation, language-prior
dominance, has motivated prior-suppression methods, but this explanation has
not been directly tested at the attribute level. We present \textbf{VISOR}
(Visual-Operational Remediation), a unified framework that couples
null-image-based diagnosis with routed remediation. Its VSNR diagnostic
decomposes each prediction into a visual logit signal and a language-prior
signal. Across 10,791 negative-ground-truth samples from three VLM families and
three attribute types, the visual signal strongly predicts false positives,
whereas the language-prior signal is near chance. VISOR uses this diagnosis to
separate two failure modes: low-margin but directionally correct visual signals
in color/state attributes, and low-SNR or misaligned visual signals in material
attributes. The same diagnosis routes each query to the appropriate operator:
calibration for threshold-placement errors, abstention for training-free
low-SNR handling, or targeted visual adaptation for material failures that
prior suppression cannot correct. Across Qwen, InternVL, and LLaVA, VISOR
reduces attribute false positives without relying on the prior-dominance
assumption.
\end{abstract}

\section{Introduction}
\label{sec:intro}

Vision-language models (VLMs)~\citep{alayrac2022flamingo,li2023blip2,liu2023llava,dai2023instructblip,zhu2023minigpt4}
have achieved strong performance on object-level
benchmarks~\citep{liu2023mmbench}, yet they remain prone to
\textit{attribute hallucination}: correctly identifying an object while
mischaracterizing its properties. A model may recognize a jacket yet assert it
is \textit{leather} rather than \textit{denim}, or describe a surface as
\textit{paper} when it is \textit{cardboard}. Unlike object-existence errors,
attribute errors can pass standard object-level checks while corrupting
decisions that depend on fine-grained visual properties.

The dominant explanation for VLM hallucination is \textbf{language prior
dominance}: statistical co-occurrences from pre-training allow language-layer
biases to override visual evidence~\citep{leng2024vcd}. This hypothesis has
motivated contrastive decoding (VCD;~\citealp{leng2024vcd}), instruction-based
correction (ICD;~\citealp{wang2024icd}), and preference
optimization~\citep{zhao2023hadpo}---all targeting the same causal pathway:
suppress the language prior, reduce hallucination. Yet whether this pathway
actually drives \textit{attribute} hallucination has never been directly tested.
This leaves a basic causal question unresolved: when an attribute false positive
occurs, is the error driven by the language prior, by the visual representation,
or by the decision boundary between them?

We address this gap with \textbf{VISOR}, a unified diagnosis-and-remediation
framework for controlled yes/no attribute probing. VISOR first decomposes each
prediction into a \textbf{visual logit difference},
$\dvis = \text{logit}(\text{yes}|\text{image}) - \text{logit}(\text{no}|\text{image})$,
and a \textbf{prior logit difference},
$\dprior = \text{logit}(\text{yes}|\text{null}) - \text{logit}(\text{no}|\text{null})$,
where the null image is a uniform blank that carries no semantic content. Across
10,791 negative-ground-truth samples on VAW~\citep{pham2021vaw}, the visual
signal is a much stronger predictor of false positives than the language-prior
signal. This finding shifts the explanation from uniform prior dominance toward
visual-signal quality, while providing a coordinate for layer-wise diagnosis and
intervention.

The same diagnostic coordinate also determines how VISOR intervenes. When the
visual signal has the correct direction but an insufficient margin, VISOR shifts
the decision boundary by a calibrated prior term. When the signal is low-SNR or
misaligned in the final Yes/No logit space, VISOR either abstains without
training or repairs the visual projection with a targeted adapter. Thus the
method is organized around one question throughout the paper: which signal
actually causes the attribute false positive, and what intervention matches that
signal?

\begin{figure}[ht]
  \centering
  \includegraphics[width=\linewidth]{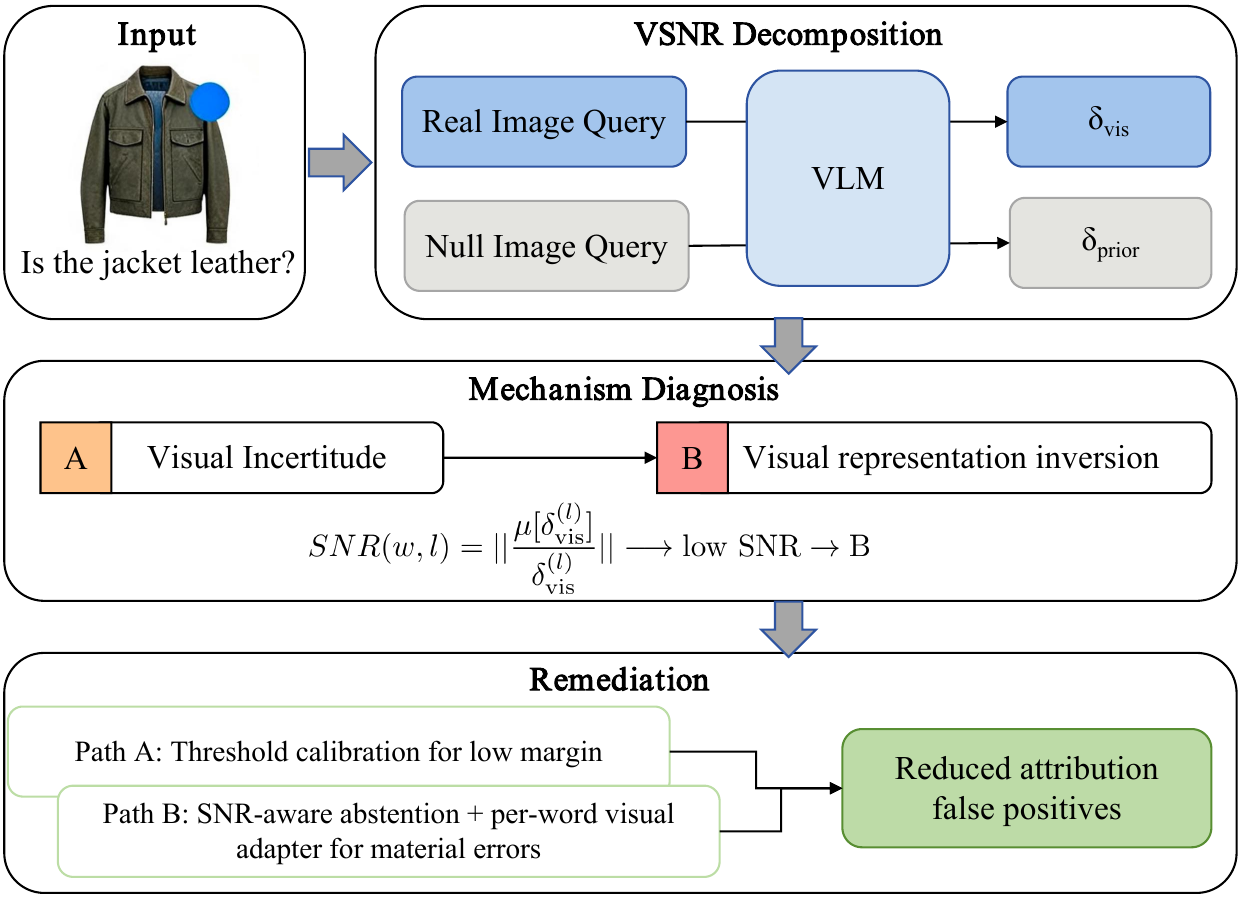}
  \caption{\textbf{Framework overview.} VISOR first uses VSNR to decompose each
    attribute query into visual and null-image prior signals. The same
    diagnostic signal supports mechanism analysis, layer-wise back-tracing, and
    routing to a mechanism-specific operator: Calib for low-margin visual
    signals, and Abstain or Adapt for low-SNR material signals.}
  \label{fig:overview}
\end{figure}

Our contributions are:
\begin{enumerate}
  \item \textbf{VISOR framework}: We introduce a unified framework that couples
    VSNR diagnosis with routed remediation, using the same visual-signal
    coordinate for hypothesis testing, layer-wise analysis, and operator
    selection.
  \item \textbf{Mechanistic taxonomy}: We identify two visual-signal failure
    modes: threshold-correctable low-margin errors in color/state and
    directional visual-signal failures in material, supported by layer-wise and
    word-level evidence.
  \item \textbf{Mechanism-specific remediation}: We evaluate VISOR as a routed
    system whose Calib, Abstain, and Adapt operators are selected by the
    diagnosed failure mode, showing that mechanism-aligned remediation
    outperforms prior-suppression baselines.
\end{enumerate}

\section{Related Work}
\label{sec:related}

\textbf{Hallucination evaluation.}
Object-level hallucination is commonly evaluated with POPE~\citep{li2023pope},
CHAIR~\citep{rohrbach2018chair}, and related benchmarks summarized in recent
surveys~\citep{yin2023vlm_survey,zhang2024vlm_survey,bai2024hallucination_survey,liu2024hallucination_survey}.
HallusionBench~\citep{liu2024hallusion}, MME~\citep{fu2023mme}, and
GQA~\citep{hudson2019gqa} include attribute-related questions, but they usually
report aggregate accuracy rather than per-attribute false positives. VAW
\citep{pham2021vaw} provides dense attribute annotations; we repurpose it as a
mechanistic yes/no diagnostic by separating false positives and false negatives
for each attribute word.

\textbf{Attribute recognition vs.\ attribute hallucination.}
ABO~\citep{collins2022abo}, OVAD~\citep{bravo2023ovad}, and
VASR~\citep{bitton2023vasr} evaluate attribute recognition with region proposals
or task-specific models. Our setting instead asks why a full VLM asserts absent
attributes during zero-shot inference, and connects per-word representation
quality to false positive rates without privileged region annotations.

\textbf{Prior-suppression interventions.}
VCD~\citep{leng2024vcd}, ICD~\citep{wang2024icd},
DAMRO~\citep{gong2024damro}, DeCo~\citep{yao2024deco}, and related contrastive
decoding methods suppress outputs attributed to language or degraded-visual
priors. Preference methods such as V-DPO~\citep{xie2024vdpo},
HA-DPO~\citep{zhao2023hadpo}, and RLHF-V~\citep{yu2024halva} align responses
toward visually faithful answers. These methods are well matched to failures
where a correct visual signal is overridden; our analysis identifies material
attribute cases where the visual signal itself is low-SNR or directionally
misaligned, requiring different interventions.

\section{VISOR: Diagnosis and Mechanism-Specific Remediation}
\label{sec:framework}

VISOR couples a diagnostic module with mechanism-specific remediation. The
diagnostic module, VSNR, decomposes each attribute prediction into two scalar
quantities extracted by paired real-image and null-image forward passes. These
quantities define the decision variable, the prior estimate, and the coordinate
used for layer-wise back-tracing. VISOR then routes a query to one of three
operators according to the diagnosed failure mode: Calib for low-margin visual
signals, Abstain for training-free low-SNR handling, and Adapt for targeted
visual repair. The operators are therefore not separate methods; they are
different actions taken by one diagnosis-driven framework.

\textbf{Formulation.} Let $M$ be a VLM, $I$ an image, and $q_w$ a yes/no attribute query ``Is the
[object] [attribute $w$]?''. We define the \textbf{visual logit difference}:
\begin{equation}
  \dvis(M, I, w) = \text{logit}_M(\text{yes} \mid I, q_w) - \text{logit}_M(\text{no} \mid I, q_w)
  \label{eq:dvis}
\end{equation}
For a negative sample (ground truth: No), hallucination occurs when $\dvis \geq 0$.

We additionally define the \textbf{prior logit difference} under a blank image
$I_\varnothing$ (uniform gray):
\begin{equation}
  \dprior(M, w) = \text{logit}_M(\text{yes} \mid I_\varnothing, q_w) - \text{logit}_M(\text{no} \mid I_\varnothing, q_w)
  \label{eq:dprior}
\end{equation}
$\dprior$ measures the model's language-only bias for word $w$ independent of
visual input. We collect both quantities for every sample via two forward passes
per (image, query) pair. We use uniform gray ($128 \times 128 \times 3$, all
channels $= 128$) as the null image; supplementary validation summarizes the
remaining scope limits of this choice.

\textbf{Decision and layer-wise coordinates.} Under deterministic yes/no decoding, the first answer token fixes the label, so
$\dvis=0$ is the decision boundary for negative samples. Empirically,
$\text{sign}(\dvis)$ matches the hallucination label across all 10,791
negative-ground-truth samples. Because the same final norm and LM head can be
applied to intermediate hidden states, we also define:
\begin{equation}
  \begin{aligned}
  \dvis^{(l)} &= z_l(\text{yes}) - z_l(\text{no}),\\
  z_l(t) &= \text{LMHead}(\text{Norm}(h_l^{[\text{last}]}))\big|_t .
  \end{aligned}
  \label{eq:dvis_layer}
\end{equation}
This gives a per-layer signal-to-noise ratio,
$\snr(w,l)=|\mu[\dvis^{(l)}]|/\sigma[\dvis^{(l)}]$, allowing us to trace when the
decision-relevant visual signal becomes low-margin or misaligned.

\textbf{Bayesian decomposition and the CD-method failure mode.}
At the final token position, the model's log-odds can be written as:
\begin{equation}
  \log \frac{P(\text{yes} \mid I, q_w)}{P(\text{no} \mid I, q_w)}
  = \underbrace{\log \frac{P(I \mid \text{yes}, q_w)}{P(I \mid \text{no}, q_w)}}_{\text{visual likelihood ratio}}
  + \underbrace{\log \frac{P(\text{yes} \mid q_w)}{P(\text{no} \mid q_w)}}_{\text{language prior}}
  \label{eq:bayes}
\end{equation}
In the limit where $I=I_\varnothing$ carries no semantic content, the expression
reduces to $\dprior$, and $\dvis-\dprior$ approximates the visual likelihood
contribution. This decomposition clarifies when prior-suppression methods are
misaligned: subtracting a prior estimate helps only if the visual signal is
directionally correct but underweighted. For material B2 words, the measured
prior estimate is non-positive while $\dvis>0$ on hallucinating samples, so
output-level prior subtraction cannot flip the decision.

For an output-level contrastive correction
$\delta'_{\mathrm{vis}}=\dvis-\alpha\hat{\delta}_{\mathrm{prior}}$, correcting a
hallucinating sample with $\dvis>0$ requires
$\hat{\delta}_{\mathrm{prior}}>\dvis/\alpha>0$. The tested B2 material words
instead have $\hat{\delta}_{\mathrm{prior}}\leq0$ across models; therefore
output-level prior subtraction cannot flip the decision sign. The formal
statement and proof are given in the supplementary material.

\textbf{Unified routing.}
VISOR uses the same diagnostic coordinate for both analysis and intervention.
For words whose visual signal is directionally correct but low-margin, VISOR
applies a calibrated decision rule
$\mathbf{1}[\dvis-\gamma\dprior>0]$. For low-SNR material words, the routed
operator depends on deployment constraints: VISOR-Abstain returns uncertainty
without training, whereas VISOR-Adapt loads a per-word visual LoRA adapter when
definitive answers are required. This routing view is important: Calib, Abstain,
and Adapt are not stitched together post hoc, but are the three actions exposed
by the same VSNR diagnosis.

\section{Experiments}
\label{sec:experiments}

\subsection{Experimental setup.}
\label{sec:setup}
We use the VAW (Visual Attributes in the Wild) dataset~\citep{pham2021vaw},
which provides per-instance attribute annotations from Visual Genome across 620
attribute words. For each attribute type we construct a yes/no probing benchmark
by issuing queries of the form \textit{``Is the [object] [attribute]? Answer yes
or no.''} Negative samples are drawn from instances where the attribute appears
in \texttt{negative\_attributes}; positive samples from
\texttt{positive\_attributes}. We measure \textbf{False Positive Rate (FPR)}---the
fraction of negative-ground-truth samples where the model answers ``yes''---as
the primary hallucination metric. Dataset statistics (Qwen2.5-VL-3B): color
2,992 samples (1,302 pos / 1,690 neg, 61 words); material 2,417 samples (969
pos / 1,448 neg, 45 words); state 803 samples (344 pos / 459 neg, 11 words). We
evaluate Qwen2.5-VL-3B-Instruct~\citep{bai2025qwen25vl},
InternVL3.5-4B-Flash~\citep{chen2024internvl25}, and
LLaVA-1.5-7B~\citep{liu2024llava15}. These three families span different visual
encoder designs (Qwen-VL ViT, InternViT-300M, CLIP ViT-L/14~\citep{radford2021clip,dosovitskiy2021vit}) and training
recipes, providing architectural diversity for cross-model validation.

\textbf{Logit extraction.}
Each (image, query) pair receives two forward passes: one with the real image to
obtain $\dvis$, one with a blank gray image ($128 \times 128 \times 3$, all
channels $= 128$) to obtain $\dprior$. We record logit(yes) and logit(no) as the
maximum log-probability among surface forms \{``Yes'', ``yes'', `` Yes'', `` yes''\}
and \{``No'', ``no'', `` No'', `` no''\} respectively from the final token position.

\textbf{Metrics.}
Our yes/no probing benchmark follows the POPE~\citep{li2023pope} format:
balanced positive/negative queries over a closed vocabulary. \FPR{} is the
primary metric because it isolates the hallucination direction; Yes-Ratio, F1,
and Accuracy provide complementary POPE-style evaluation.

\begin{table*}[ht]
\centering
\small
\caption{\textbf{Method comparison on the VAW probing benchmark (F1 / Accuracy /
Yes-Ratio / FPR).} Best \FPR{} per column is in \textbf{bold}. VISOR denotes
the routed framework: Calib is selected for color/state and Adapt for material.
VCD/ICD Yes-Ratio increases reveal worsening yes-bias masked by stable
F1/Accuracy.}
\label{tab:method_comparison}
\smallskip
\textit{(a) Color attribute}
\resizebox{\linewidth}{!}{%
\begin{tabular}{lcccccccccccc}
\toprule
 & \multicolumn{4}{c}{Qwen2.5-VL-3B} & \multicolumn{4}{c}{InternVL3.5-4B} & \multicolumn{4}{c}{LLaVA-1.5-7B} \\
\cmidrule(lr){2-5}\cmidrule(lr){6-9}\cmidrule(lr){10-13}
Method & F1 & Acc & YR & FPR & F1 & Acc & YR & FPR & F1 & Acc & YR & FPR \\
\midrule
Baseline       & 73.7 & 73.9 & 55.6 & 33.8 & 74.7 & 75.2 & 54.4 & 31.6 & 67.8 & 65.8 & 62.6 & 47.2 \\
VCD            & 73.7 & 72.9 & 59.4 & 38.0 & 74.9 & 74.7 & 57.2 & 35.6 & 68.1 & 65.4 & 64.8 & 51.3 \\
ICD            & 73.9 & 73.4 & 58.2 & 36.5 & 74.7 & 74.5 & 57.6 & 35.6 & 67.9 & 64.9 & 66.1 & 51.3 \\
\textbf{VISOR (ours)} & 71.4 & \textbf{75.1} & \textbf{43.7} & \textbf{22.2} & 71.7 & \textbf{76.9} & \textbf{37.8} & \textbf{15.4} & 65.2 & \textbf{68.9} & \textbf{46.1} & \textbf{29.8} \\
\bottomrule
\end{tabular}
}
\smallskip
\textit{(b) Material attribute}
\resizebox{\linewidth}{!}{%
\begin{tabular}{lcccccccccccc}
\toprule
 & \multicolumn{4}{c}{Qwen2.5-VL-3B} & \multicolumn{4}{c}{InternVL3.5-4B} & \multicolumn{4}{c}{LLaVA-1.5-7B} \\
\cmidrule(lr){2-5}\cmidrule(lr){6-9}\cmidrule(lr){10-13}
Method & F1 & Acc & YR & FPR & F1 & Acc & YR & FPR & F1 & Acc & YR & FPR \\
\midrule
Baseline              & 73.0 & 79.0 & 37.9 & 15.7 & 74.2 & 78.8 & 42.3 & 19.5 & 70.5 & 74.7 & 45.7 & 25.8 \\
VCD                   & 73.7 & 78.1 & 43.4 & 15.7 & 74.8 & 78.7 & 44.2 & 19.5 & 71.3 & 74.6 & 48.5 & 25.8 \\
ICD                   & 74.4 & 78.0 & 46.0 & 15.7 & 75.0 & 78.0 & 47.7 & 19.5 & 70.7 & 75.7 & 42.9 & 25.8 \\
\textbf{VISOR (ours)} & \textbf{73.9} & \textbf{79.9} & \textbf{37.0} & \textbf{14.2} & \textbf{75.9} & \textbf{80.3} & \textbf{41.7} & \textbf{17.8} & \textbf{72.3} & \textbf{76.2} & \textbf{45.7} & \textbf{24.5} \\
\bottomrule
\end{tabular}
}
\smallskip
\textit{(c) State attribute}
\resizebox{\linewidth}{!}{%
\begin{tabular}{lcccccccccccc}
\toprule
 & \multicolumn{4}{c}{Qwen2.5-VL-3B} & \multicolumn{4}{c}{InternVL3.5-4B} & \multicolumn{4}{c}{LLaVA-1.5-7B} \\
\cmidrule(lr){2-5}\cmidrule(lr){6-9}\cmidrule(lr){10-13}
Method & F1 & Acc & YR & FPR & F1 & Acc & YR & FPR & F1 & Acc & YR & FPR \\
\midrule
Baseline       & 71.0 & 74.1 & 46.6 & 25.9 & 66.7 & 71.2 & 43.5 & 25.7 & 67.8 & 70.1 & 50.1 & 32.5 \\
VCD            & 71.0 & 73.8 & 46.0 & 25.7 & 64.8 & 67.7 & 42.7 & 28.1 & 63.5 & 62.4 & 47.5 & 37.0 \\
ICD            & 71.0 & 73.8 & 46.0 & 25.7 & 64.8 & 67.7 & 42.7 & 28.1 & 63.5 & 62.4 & 47.5 & 37.0 \\
\textbf{VISOR (ours)} & 70.4 & \textbf{75.5} & \textbf{40.0} & \textbf{19.0} & 65.8 & \textbf{72.5} & \textbf{37.7} & \textbf{19.6} & 67.8 & 70.1 & 50.1 & 32.5 \\
\bottomrule
\end{tabular}
}
\end{table*}

\textbf{Reproducibility.}
All inference experiments use greedy decoding with temperature$=0$ and fixed random seed 42. LoRA fine-tuning uses seed 42 across all three model families. Experiments were conducted on a single NVIDIA A100 80GB GPU.

\textbf{Language prior does not drive attribute hallucination.} We assess whether the language prior ($\dprior$) or visual signal ($\dvis$)
explains attribute hallucination. We present a primary correlational analysis
followed by an additional image-sensitivity analysis.

\textbf{Correlational analysis.}
For 10,791 negative-ground-truth samples, we compute Spearman rank correlation
between each logit quantity and the binary hallucination outcome
(1 = model predicted ``yes''):

\begin{table}[ht]
\centering
\small
\caption{Spearman rank correlation between logit signals and binary hallucination outcome. $\dvis$ strongly predicts hallucination across all three models; $\dprior$ is near chance. A classifier using $\dprior$ alone achieves 58.5--59.0\% accuracy---barely above the majority-class baseline (57.9\%).}
\resizebox{\linewidth}{!}{%
\begin{tabular}{lccc}
\toprule
Model & $\rho(\dvis, \text{halluc.})$ & $\rho(\dprior, \text{halluc.})$ & Prior-only acc. \\
\midrule
Qwen2.5-VL-3B & \textbf{0.755} & 0.177 & 58.5\% \\
InternVL3.5-4B & \textbf{0.760} & 0.066 & 58.6\% \\
LLaVA-1.5-7B & \textbf{0.835} & 0.066 & 59.0\% \\
\bottomrule
\end{tabular}
}
\label{tab:correlation}
\end{table}

This result should be read as a scoped claim. It does not deny language bias in general: prior work documents attestation and frequency biases in language models~\citep{mckenna2023hallucination}, and VLMs can retain substantial benchmark accuracy without images~\citep{asadi2025mirage}. It also does not claim that $\dvis$ is independent of language pre-training, since visual encoders and language heads are jointly trained. The narrower finding is that, at inference time in negative-ground-truth attribute probing, the visual logit margin explains sample-level false positives better than the null-image prior.

\textbf{Image-sensitivity analysis (additional evidence).}
We next test whether the real image can override a positive null-image prior. We identify 278 \textit{prior-positive conflict} cases where $\dprior > 0$ but $\dvis < 0$:

\begin{table}[ht]
\centering
\small
\caption{Prior-positive conflict cases ($\dprior>0$, $\dvis<0$). In all 278
cases where blank-image prediction favors ``yes'' but real image flips the
prediction to ``no'', no hallucination occurs, confirming that real image content
actively modifies model behavior.}
\resizebox{\linewidth}{!}{%
\begin{tabular}{lcc}
\toprule
Model & Prior-pos.\ conflicts & Halluc.\ rate \\
\midrule
Qwen2.5-VL-3B & 147 / 1,690 & \textbf{0.0\%} \\
InternVL3.5-4B & 74 / 1,690 & \textbf{0.0\%} \\
LLaVA-1.5-7B & 57 / 1,690 & \textbf{0.0\%} \\
\bottomrule
\end{tabular}
}
\label{tab:conflict}
\end{table}

\textbf{Language prior predicts color FPR but not material FPR.}
At the word level, $\dprior$ provides asymmetric diagnostic information. For color attributes, the per-word mean $\dprior$ correlates with FPR at $\rho = +0.420$ ($p = 0.003$, $n = 49$ words). For material attributes,
$\rho(\dprior, \FPR) = +0.034$ ($p = 0.83$)---no relationship. This asymmetry
explains why VISOR-Calib succeeds for color/state but achieves zero improvement
on material: for material, word-level FPR variance is explained entirely by
$\dvis$ distribution properties ($\rho = +0.752$, $p < 0.0001$).

\section{Two Mechanisms of Attribute Hallucination}
\label{sec:mechanisms}

\textbf{Mechanism A: Visual Incertitude.}
For \textbf{color} and \textbf{state} attributes, $\dvis$ points in the correct
direction on average but with insufficient margin. VISOR-Calib
reduces color FPR by 11--17~pp across all three models and reduces state FPR by
6--7~pp on Qwen and InternVL, while VCD/ICD achieve 0.0~pp improvement on color.
This pattern supports a threshold-misplacement account rather than a
prior-dominance account. Compound-color diagnostics are provided in the
supplementary material.

\textbf{Mechanism B: Visual Representation Inversion.}
For \textbf{material} attributes, high FPR persists even for words with near-zero
language prior. We separate annotation granularity noise from true representation
failure. Steel is mostly B1 label granularity: 18/20 hallucinating negatives are
visually metallic, so its true FPR drops from 45.5\% to 4.5\% after removing
annotation-noise cases. Rubber, paper, and leather are B2 failures: observed and
true FPR are identical after the same filtering.

\begin{table}[ht]
\centering
\small
\caption{Observed vs. true FPR after annotation-noise removal. B1 words show lower true FPR; B2 words remain unchanged, confirming genuine visual representation failure.}
\label{tab:b1b2}
\resizebox{\linewidth}{!}{%
\begin{tabular}{lccl}
\toprule
Word & Observed FPR & True FPR & Type \\
\midrule
steel & 45.5\% & 4.5\% & B1 \\
rubber & 44.4\% & 44.4\% & B2 \\
paper & 41.9\% & 41.9\% & B2 \\
ceramic & 32.5\% & 27.5\% & mixed \\
leather & 26.0\% & 26.0\% & B2 \\
\bottomrule
\end{tabular}
}
\end{table}

\textbf{SNR predicts material severity.}
We define $\snr(w)=|\bar{\delta}_{\text{vis}}(w)|/\sigma_{\delta_{\text{vis}}(w)}$.
Across 46 material words, SNR is the strongest predictor of FPR.

\begin{table}[ht]
\centering
\small
\caption{Predictors of material FPR across 46 material words. SNR achieves $\rho=-0.916$, outperforming either component alone.}
\label{tab:snr_predictors}
\resizebox{\linewidth}{!}{%
\begin{tabular}{lcc}
\toprule
Predictor & Spearman $\rho$ with FPR & $p$-value \\
\midrule
$\bar{\delta}_{\text{vis}}(w)$ & $-0.765$ & $< 0.0001$ \\
$\sigma_{\delta_{\text{vis}}(w)}$ & $+0.633$ & $< 0.0001$ \\
$\snr(w)$ & $-0.916$ & $< 0.0001$ \\
\bottomrule
\end{tabular}
}
\end{table}

\begin{figure}[ht]
  \centering
  \includegraphics[width=\linewidth]{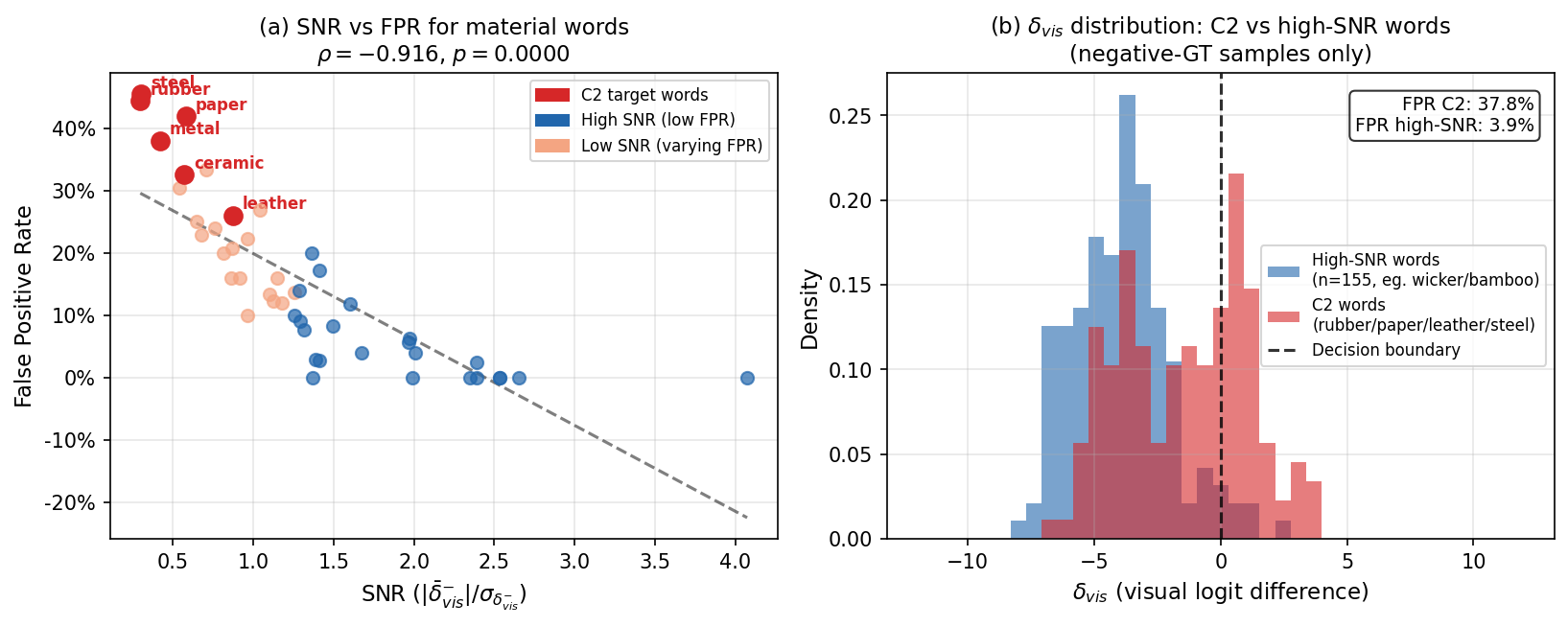}
  \caption{SNR vs. FPR and $\dvis$ distributions for material words. B2 words have broad $\dvis$ distributions straddling zero, while high-SNR words concentrate below zero.}
  \label{fig:snr_feature}
\end{figure}

\textbf{Late-stage SNR collapse.}
Layer-wise probing localizes Mechanism~B to the final decoder layers rather than
to a complete absence of mid-layer visual evidence. For B2 words, SNR drops from
about 35 near L17 to below 0.9 by L36, whereas high-SNR words maintain SNR around
2--4. This narrows the failure to late-stage visual-to-logit projection dynamics;
cross-model trajectories, signed-signal plots, and case studies are in the
supplementary material.

\begin{figure}[ht]
  \centering
  \includegraphics[width=\linewidth]{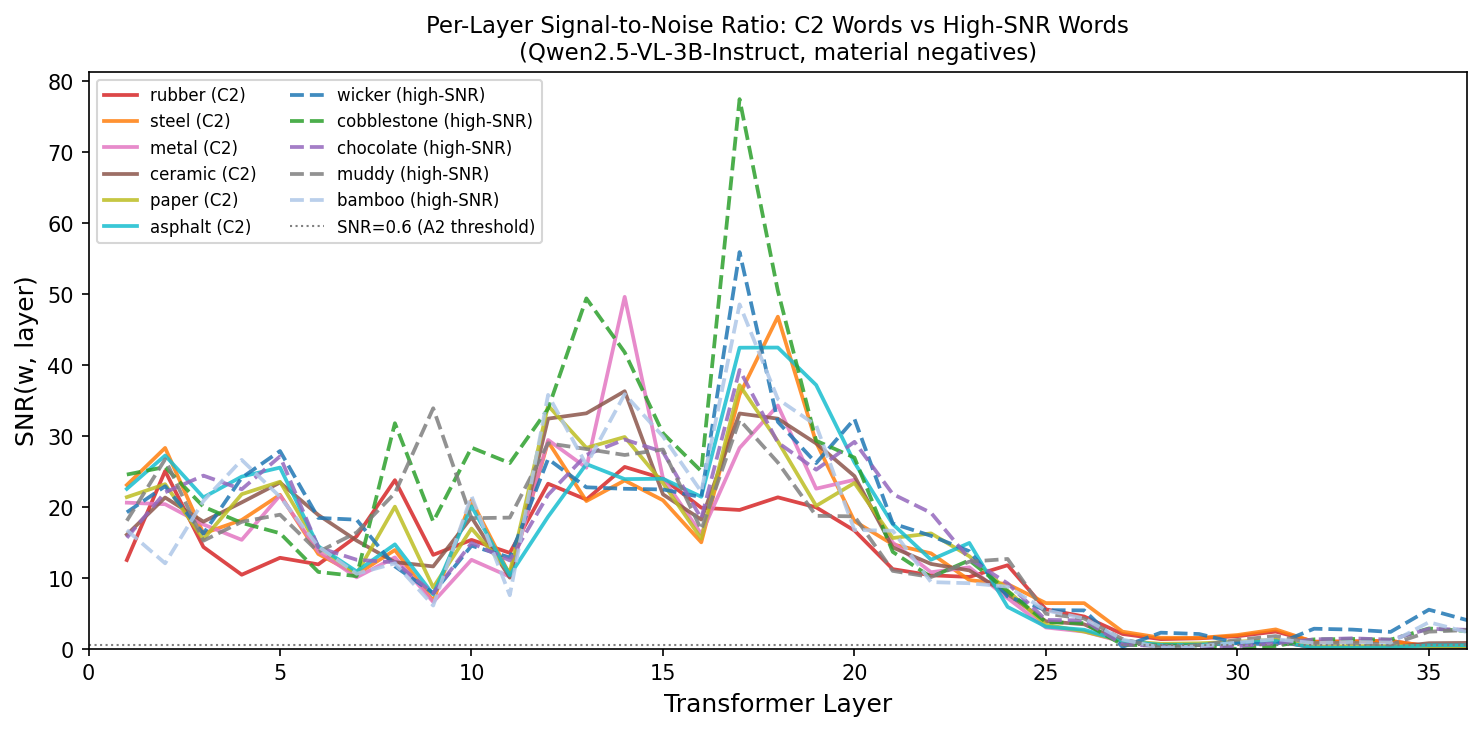}
  \caption{Per-layer SNR for low-SNR material words vs. high-SNR words across 36 decoder layers of Qwen2.5-VL-3B. The divergence occurs in L28--L36, localizing the degradation to late decoder layers.}
  \label{fig:layer_snr}
\end{figure}

\section{VISOR Results and Mechanistic Validation}
\label{sec:remediation}

We next verify that VISOR's routed operators match the diagnosed mechanisms.
These experiments should be read as branch-level validation of one framework,
not as a collection of unrelated fixes. For Mechanism~A, where $\dvis$ is
directionally correct but low-margin, the Calib operator shifts the decision
boundary using the null-image prior and reduces color FPR by 11--17~pp across
Qwen, InternVL, and LLaVA. For material attributes, the same operator gives
0~pp on Qwen and InternVL, matching the diagnosis that Mechanism~B is not
threshold correctable.

\begin{table}[ht]
\centering
\small
\caption{VISOR Calib operator vs.\ VCD/ICD on color, state, and material across
three models. Calib consistently helps color/state when $\dvis$ is directionally correct but
low-margin; its zero-gain material results on Qwen/InternVL confirm that
Mechanism~B is not corrected by threshold movement. For LLaVA material, Calib
reduces FPR by shifting the threshold but increases FNR by 7.1~pp.}
\label{tab:path_a}
\resizebox{\linewidth}{!}{%
\begin{tabular}{llcccc}
\toprule
Model & Attribute & Baseline & VISOR-Calib & VCD & ICD \\
\midrule
Qwen2.5-VL-3B & color & 33.8\% & \textbf{22.2\%} ($-$11.6~pp) & 33.8\% (0~pp) & 33.8\% (0~pp) \\
InternVL3.5-4B & color & 31.6\% & \textbf{15.4\%} ($-$16.2~pp) & 31.6\% (0~pp) & 31.6\% (0~pp) \\
LLaVA-1.5-7B & color & 47.2\% & \textbf{29.8\%} ($-$17.4~pp) & 47.2\% (0~pp) & 47.2\% (0~pp) \\
\midrule
Qwen2.5-VL-3B & state & 25.9\% & \textbf{18.9\%} ($-$7.0~pp) & 25.7\% (0~pp) & 25.7\% (0~pp) \\
InternVL3.5-4B & state & 25.7\% & \textbf{19.6\%} ($-$6.1~pp) & 28.1\% ($+$2.4~pp) & 28.1\% ($+$2.4~pp) \\
LLaVA-1.5-7B & state & 32.5\% & 32.5\% (\textbf{0.0~pp}) & 37.0\% ($+$4.6~pp) & 37.0\% ($+$4.6~pp) \\
\midrule
Qwen2.5-VL-3B & material & 15.7\% & 15.7\% (\textbf{0.0~pp}) & 15.7\% (0~pp) & 15.7\% (0~pp) \\
InternVL3.5-4B & material & 19.5\% & 19.5\% (\textbf{0.0~pp}) & 19.5\% (0~pp) & 19.5\% (0~pp) \\
LLaVA-1.5-7B & material & 25.8\% & \textbf{18.6\%} ($-$7.2~pp) & 25.8\% (0~pp) & 25.8\% (0~pp) \\
\bottomrule
\end{tabular}
}
\end{table}

\begin{figure}[ht]
  \centering
  \includegraphics[width=\linewidth]{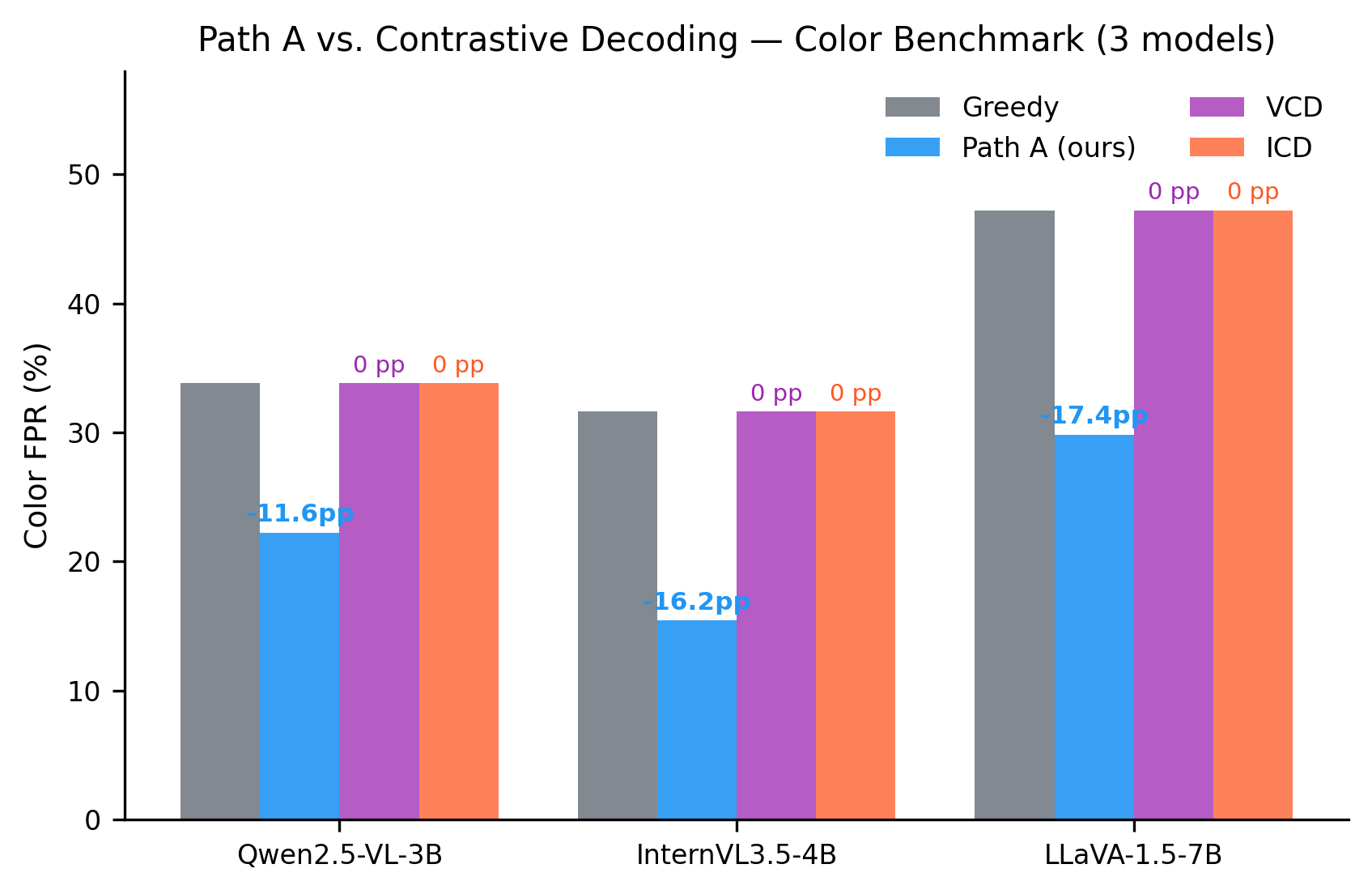}
  \caption{The VISOR Calib operator reduces color FPR across Qwen, InternVL, and LLaVA, whereas
    VCD and ICD do not improve color FPR. This supports the Mechanism~A diagnosis:
    the relevant error is boundary placement under a weak but correct visual
    signal, not a dominant language prior that can be subtracted away.}
  \label{fig:path_a}
\end{figure}

For Mechanism~B, output-level prior subtraction and language-layer tuning are
insufficient because the discriminative material signal is weak or poorly
projected. The VISOR Abstain operator, a training-free routed action, reduces Qwen material FPR
from 15.7\% to 10.2\% at $\tau=0.6$ with 14.2\% abstention. More direct repair
uses the Adapt operator: a shared hard-negative LoRA suffers cross-contamination,
whereas Adapt trains one visual LoRA adapter per high-FPR target word. On
the six target material words, VISOR-Adapt reduces mean FPR from 38.1\% to
23.9\% on Qwen, 44.7\% to 32.8\% on InternVL, and 34.2\% to 26.8\% on LLaVA,
with MME regression within $\pm$1.25~pp. Full probe results, per-word tables,
statistical tests, and free-form transfer results are in the supplementary
material.

\begin{table}[ht]
\centering
\small
\caption{VISOR-Abstain on Qwen2.5-VL-3B material queries. The
$\tau=0.6$ operating point lowers FPR by 5.5~pp with 14.2\% abstention, showing a
training-free option for low-SNR material words; $\tau=1.0$ is more conservative
but abstains on nearly half of the samples.}
\label{tab:a2_abstention}
\resizebox{\linewidth}{!}{%
\begin{tabular}{lccc}
\toprule
$\tau$ & FPR & $\Delta$ FPR & Abstain rate \\
\midrule
0 (baseline) & 15.7\% & 0.0~pp & 0.0\% \\
0.3 & 14.1\% & $-$1.6~pp & 6.2\% \\
0.6 & \textbf{10.2\%} & \textbf{$-$5.5~pp} & 14.2\% \\
1.0 & 4.8\% & $-$10.9~pp & 45.5\% \\
\bottomrule
\end{tabular}
}
\end{table}

\begin{figure}[ht]
  \centering
  \includegraphics[width=\linewidth]{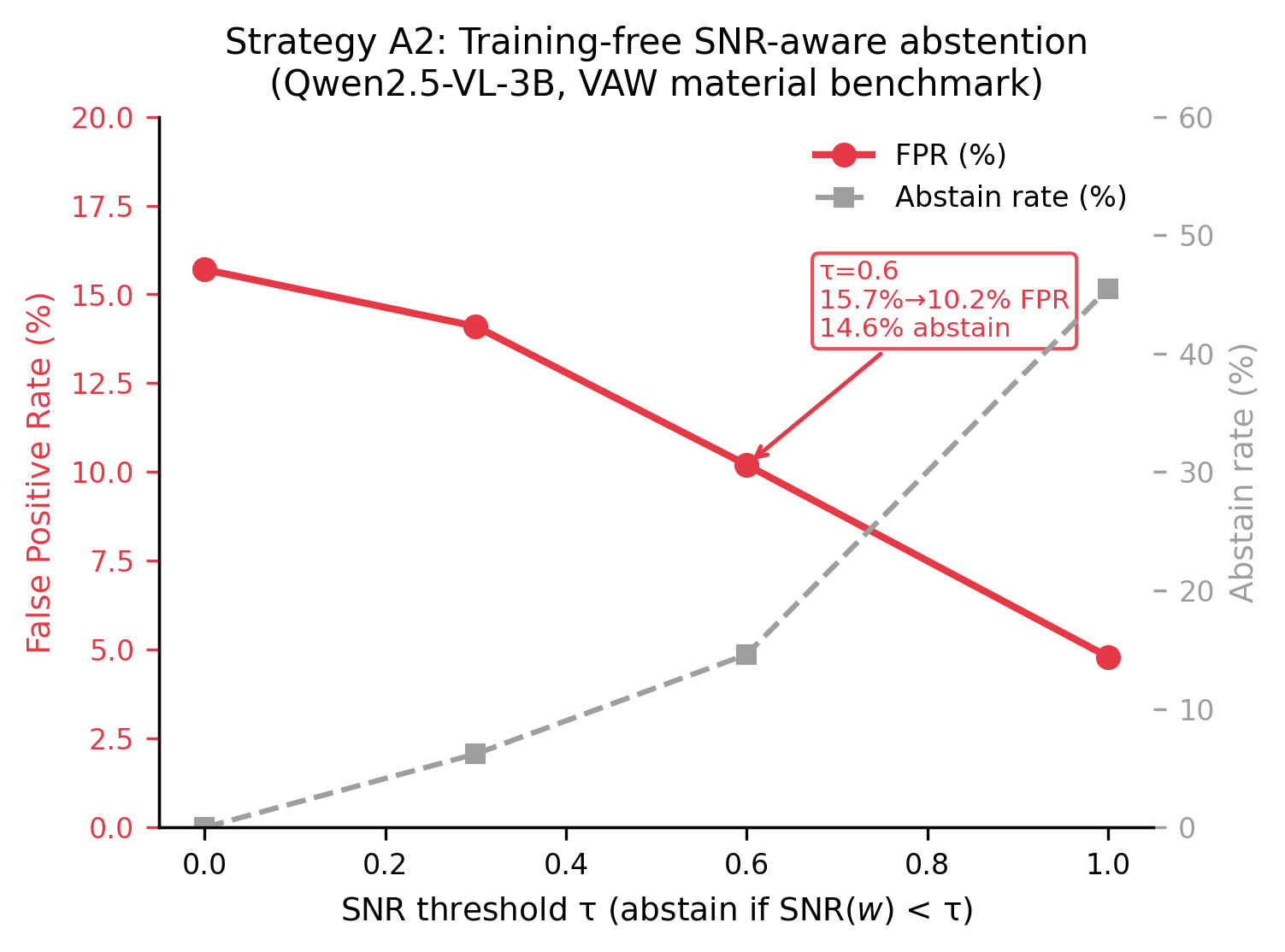}
  \caption{VISOR-Abstain traces the FPR--coverage trade-off as the SNR threshold increases.
    Abstention concentrates on the highest-risk material words, making the
    intervention appropriate for safety-sensitive settings where unsupported
    material assertions are costlier than uncertainty.}
  \label{fig:a2_abstention}
\end{figure}

\textbf{VISOR-Adapt repairs material-specific visual projections.}
VISOR-Adapt trains one visual LoRA adapter per high-FPR material word, avoiding the
cross-contamination observed with a shared hard-negative LoRA. On Qwen2.5-VL-3B,
the six-word mean FPR drops from 38.1\% to 23.9\%; the leather target illustrates
the difference most clearly, where shared LoRA worsens FPR by 8~pp but
VISOR-Adapt reduces it by 12~pp.

\begin{table}[ht]
\centering
\small
\caption{Per-word material FPR for Qwen2.5-VL-3B under baseline, shared
hard-negative LoRA, and VISOR-Adapt per-word LoRA. VISOR-Adapt improves all six
targets and avoids the leather cross-contamination failure of the shared
adapter.}
\label{tab:b3_qwen}
\resizebox{\linewidth}{!}{%
\begin{tabular}{lcccc}
\toprule
Word & Baseline & HN (shared) & VISOR-Adapt & 95\% CI \\
\midrule
steel & 45.5\% & 27.3\% & \textbf{22.7\%} & [10.3\%, 35.1\%] \\
rubber & 44.4\% & 33.3\% & \textbf{27.8\%} & [14.2\%, 41.4\%] \\
paper & 41.9\% & 38.7\% & \textbf{19.4\%} & [9.5\%, 29.3\%] \\
metal & 38.0\% & 26.0\% & 32.0\% & [19.2\%, 44.8\%] \\
leather & 26.0\% & 34.0\%~$\uparrow$ & \textbf{14.0\%}~$\downarrow$ & [5.1\%, 22.9\%] \\
ceramic & 32.5\% & 27.5\% & \textbf{27.5\%} & [15.8\%, 39.2\%] \\
\midrule
\textbf{Mean (6)} & \textbf{38.1\%} & 31.1\% & \textbf{23.9\%} & --- \\
\bottomrule
\end{tabular}
}
\end{table}

The late-stage overwriting account predicts that VISOR-Adapt should improve final-layer
material SNR rather than merely suppressing ``yes'' answers. Re-running the
layer-wise probe after VISOR-Adapt training confirms this prediction: final-layer SNR
increases for rubber (+0.307), steel (+0.407), metal (+0.177), ceramic (+0.246),
and paper (+0.256), with the gains concentrated in L28--L36.

\begin{figure}[ht]
  \centering
  \includegraphics[width=\linewidth]{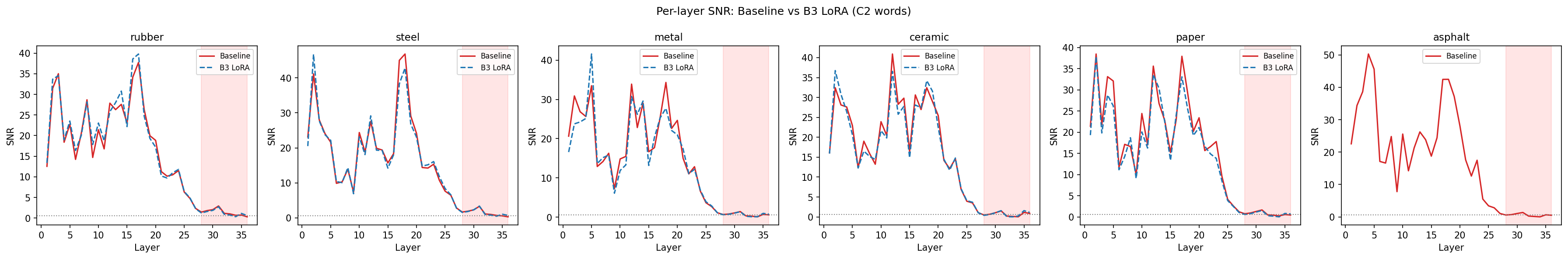}
  \caption{VISOR-Adapt raises final-layer SNR for five B2 material words on
    Qwen2.5-VL-3B. The improvement is concentrated in L28--L36, matching the
    late-stage projection failure identified by the VSNR diagnostic and
    supporting VISOR-Adapt as a mechanism-aligned visual adaptation rather than
    a generic response suppressor.}
  \label{fig:b3_snr}
\end{figure}

\textbf{Scope and limitations.}
VISOR's VSNR diagnostic is designed for structured attribute queries over known words.
Null-image estimates are stable across gray, black, white, and random-noise
images ($r>0.90$), but the binary yes/no decomposition may not capture all
free-form hallucination phenomena. B2 errors may partly reflect joint
language-visual co-training frequency rather than purely visual encoder failure:
the L28 GT probe suggests that some material information remains recoverable
before final projection, but our experiments do not separate pre-training
coverage from decoder projection. VISOR-Adapt currently targets six high-FPR
material words (16.1\% of material test samples), so it supports targeted
repair rather than open-vocabulary material coverage.

\section{Discussion}
\label{sec:discussion}

The main empirical implication is not that language statistics are irrelevant
to VLM behavior, but that attribute false positives should not be treated as a
single prior-dominance failure. VISOR separates the decision margin observed
under the real image from the null-image prior and uses this separation to pick
the intervention. This matters because methods that look similar at the output
layer can have different effects depending on the diagnosed failure mode: Calib
is useful when the visual signal has the right sign but an overly permissive
threshold, whereas prior subtraction and language-layer tuning have little room
to help when the final visual logit direction is already wrong.

\begin{table}[ht]
\centering
\small
\caption{Boundary tests for Mechanism~B on Qwen2.5-VL-3B material queries.
Language-layer tuning and undifferentiated visual updates fail to repair the
failure mode; a supervised L28 probe recovers part of the material signal,
supporting the need for targeted visual adaptation.}
\label{tab:mechanism_b_boundary}
\resizebox{\linewidth}{!}{%
\begin{tabular}{lccp{3.1cm}}
\toprule
Test & FPR change & FNR change & Diagnosis \\
\midrule
Language-layer DPO & $+2.0$ to $+2.1$~pp & -- & No update reaches the visual encoder \\
Shared visual LoRA & large drop & $+100$~pp & Degenerates to always predicting ``No'' \\
Inference-time edits & $-1.7$ to $+60$~pp & unstable & Do not isolate the GT material direction \\
GT probe at L28 & \textbf{$-10.8$~pp} & $+5.6$~pp & Recoverable signal is poorly projected by lm\_head \\
\bottomrule
\end{tabular}
}
\end{table}

Table~\ref{tab:mechanism_b_boundary} summarizes the negative controls behind
the routing decision. Their value is diagnostic rather than competitive:
language-layer DPO leaves the visual-encoder gradient at zero in our frozen
visual-backbone setup, shared visual LoRA collapses to a blanket ``No'' policy,
and simple hidden-state edits are unstable. The only partial recovery comes from
a supervised L28 probe, indicating that some material evidence is present before
the final projection but is not aligned with the model's Yes/No head. This is
why VISOR-Adapt targets visual projections per word instead of applying a shared
output-level correction.

The same mechanism-aligned design transfers across architectures. On InternVL3.5-4B-Flash, VISOR-Adapt reduces the six-word mean FPR from 44.7\% to 32.8\% and improves 5 of 6 target words; on LLaVA-1.5-7B, it reduces the mean from 34.2\% to 26.8\% and improves all 6 words. These gains are statistically supported by the stratified CMH tests reported in the supplementary material, which remain significant across the three models.

The adapters also do not trade off general capability. When each per-word LoRA is loaded individually on Qwen2.5-VL-3B and evaluated on the MME hallucination subset, Accuracy stays within ±1.25 pp of the baseline (90.8\% overall, 89.6\%–91.7\% range). A one-sample two-sided t-test across the six adapter-level Accuracy differences yields p = 0.62; VISOR-Adapt repairs the targeted material words without measurably degrading unrelated visual understanding.

The diagnostic itself is also stable. The gray null image remains highly consistent with black, white, and random-noise baselines ($r > 0.90$ in all three models), and the text-only prior aligns with gray for Qwen and InternVL. This robustness matters because the routed intervention depends on the relative ordering between the real-image margin and the null-image prior, not on a single handcrafted background choice.
We also observe the same decision rule on the GQA color subset: the $\delta_{vis}\ge 0$ criterion yields zero false-positive violations on all three models, indicating that the visual-signal diagnosis generalizes beyond the VAW probing format.

Because the six adapted words account for 16.1\% of the material negative samples, the targeted repair also translates into a measurable end-to-end gain: the estimated full-material FPR on Qwen2.5-VL-3B falls from 15.7\% to 13.5\%. The remaining gap reflects the fact that the method is intentionally focused on the highest-FPR words rather than the full open vocabulary.

This routing view also clarifies deployment choices. VISOR-Calib is the least
intrusive branch: after offline calibration of $\gamma$, inference uses the
same model and changes only the decision rule. It is therefore appropriate when
the target vocabulary is known and false positives can be reduced without
changing model parameters. VISOR-Abstain is training-free and is useful when
unsupported material assertions are costlier than uncertainty, but its benefit
comes with lower coverage. VISOR-Adapt provides definitive answers for the
highest-risk material words by updating visual-encoder projections, but the
current experiments cover a targeted vocabulary rather than all possible
attributes. Scaling it to open-vocabulary deployment would require a target-word
selection policy and adapter management, such as grouping low-SNR words or
dynamically loading adapters; these engineering strategies are not evaluated in
this paper.

The pre-training source of Mechanism~B remains an open question. The L28 GT
probe indicates that some material information is recoverable before the final
projection, while the late-layer SNR drop shows that this information is not
preserved in the final Yes/No logit coordinate. This is consistent with a
visual-to-language projection bottleneck and with insufficient fine-grained
material supervision during joint pre-training, but our experiments do not
disentangle these causes. We therefore treat VISOR as a diagnostic and repair
framework for observed attribute failures, not as a complete account of how the
failures arise during pre-training.

\section{Conclusion}
\label{sec:conclusion}

We present VISOR, a unified framework that couples null-image-based VSNR
diagnosis with routed remediation for attribute hallucination. Across controlled
yes/no probing, false positives are better predicted by the visual logit signal
than by the null-image prior, motivating a visual-signal account with two
failure modes: low-margin color/state errors and low-SNR or misaligned material
errors. The intervention results close the loop: Calib corrects threshold
placement when the visual direction is reliable, while Abstain and Adapt handle
material cases where prior suppression is insufficient. These results suggest
that hallucination mitigation should start from mechanism diagnosis rather than
uniformly suppressing a presumed language prior.

\bibliography{references}

\end{document}